\pdfoutput=1

\documentclass[11pt]{article}

\usepackage[final]{acl}

\usepackage{times}
\usepackage{latexsym}
\usepackage{algorithm}
\usepackage{algorithmicx}
\usepackage{algpseudocode}
\usepackage[T1]{fontenc}
\usepackage[utf8]{inputenc}
\usepackage{microtype}
\usepackage{inconsolata}
\usepackage{adjustbox}
\usepackage{ltablex}   

\usepackage{array,graphicx}
\usepackage{arydshln}
\usepackage{amsmath}
\usepackage{tabularray}
\usepackage{color}
\usepackage[dvipsnames]{xcolor}
\usepackage{xspace}
\usepackage{cleveref}
\usepackage{enumitem}
\usepackage{threeparttable}
\usepackage{multirow}
\usepackage{tikz}
\usepackage{subcaption}
\usetikzlibrary{positioning}
\usepackage[edges]{forest}
\usepackage{booktabs}
\usepackage{rotating}
\usepackage{pifont}
\usepackage{tabularx}
\newcommand{\cmark}{\ding{51}}

\keepXColumns
\definecolor{dip}{HTML}{D62F2F}

\title{SHREK: Scoring Hop-chains for Retrieval of Evidence over Knowledge Graphs}

\author{
 \textbf{Dylan Uno Syahfaril}\quad\textbf{Bao Gia Doan}\quad\textbf{Aditya Joshi}
\\
University of New South Wales, Sydney, Australia
\\
\texttt{dylan.unos@gmail.com, \{bao.doan1, aditya.joshi\}@unsw.edu.au}
}

\begin{document}
\maketitle
\begin{abstract}
Knowledge-graph retrieval can improve multi-hop question answering in RAG systems. However, systems such as HippoRAG 2 walk the whole graph before they rank passages, and add retrieval-time LLM filtering and a dense-retrieval fallback. We ask how far a retriever gets if it stays local and walks out from the entities the question names. We introduce Scoring Hop-chains for Retrieval of Evidence over Knowledge graphs (SHREK), a training-free retriever that traverses a locally bounded, relation-labeled OpenIE graph and uses a pretrained cross-encoder to score passage chains jointly. On the task of question answering, we compare SHREK with Dijkstra and Prize-Collecting Steiner Tree-based retrieval, as well as six past works, across seven datasets from the HippoRAG 2 benchmark suite. SHREK performs within 1.1 Token F1 points of HippoRAG 2 on average, surpasses it on PopQA by 6.6 points and HotpotQA by 1.9 points, and scores above our dense retrieval baseline on all four multi-hop datasets. SHREK achieves this without any LLM call inside the retrieval loop, without a dense fallback, and without corpus-wide graph traversal. The failure analysis of SHREK identifies two main limitations: passage-chain scoring deteriorates beyond two hops, and missing seed entities can produce empty retrievals.

\end{abstract}

\section{Introduction}
Despite their impressive capabilities in question answering, large language models (LLMs) suffer from hallucinations where models produce plausible-sounding but incorrect information within answers~\citep{ji2023survey}. Retrieval-Augmented Generation (RAG) attempts to minimise hallucination by grounding responses in external knowledge sources~\citep{lewis2020retrieval}. Knowledge Graphs (KGs), which represent facts as (subject, predicate, object) triples, have been used as the knowledge source in a RAG system (KG-RAG)~\citep{guo2024lightrag}.

Recent KG-RAG systems such as HippoRAG~2~\citep{gutierrez2025rag} are characterised by three mechanisms: (a) a global walk, where Personalized PageRank (PPR)~\citep{haveliwala2002topic} spreads over the full graph, as implemented, before any passage is ranked; (b) an LLM call inside KG retrieval that filters candidate triples, with a fallback to dense retrieval when that filter returns nothing; and (c) passage chunks mixed into the graph as nodes. Matching the query to the graph before traversal starts is common to every graph retriever (\S\ref{sec:related}), so it is not one of the three. PropRAG~\citep{proprag}, which improves on HippoRAG~2, also runs PPR over the whole graph first and only narrows to a subgraph after that. We ask what a retriever can do without that walk, expanding out from the seed entities instead. Two things make this worth asking. HippoRAG~2's filter call sits inside retrieval and reads candidate triples pulled from the graph, so moving the one LLM call per question to before retrieval, where it reads the question alone, makes retrieval a fixed procedure over the graph rather than a sequence of model judgements. And a pipeline with no dense fallback shows what the graph alone contributes, since no answer can be credited to a non-graph backup. The cost is that dropping (a) and (b) takes away the parts of HippoRAG~2 that surface a passage the question barely mentions, so a local retriever needs some other way to find the weak second passage. Our answer is to score whole chains of passages together rather than one passage at a time. We also keep the relation label\footnote{The predicate of a triple: in (\textit{Metallica}, \textit{collaborated with}, \textit{San Francisco Symphony}) the label is \textit{collaborated with}.} on every edge, which PPR ignores, and use it to prune candidates at each hop. We present \textbf{S}coring \textbf{H}op-chains for \textbf{R}etrieval of \textbf{E}vidence over \textbf{K}nowledge graphs (SHREK), a KG-RAG pipeline built without those three mechanisms. Our hypothesis is not that removing them raises accuracy. It is that a retriever without them can stay close to HippoRAG~2, and the experiments are set up to show how close it gets and where the gap opens.

SHREK retrieves supporting passages using a knowledge graph, \textit{but it does not call an LLM inside the retrieval loop and does not fall back to dense retrieval}, which removes mechanism (b). It makes one LLM call on the question before retrieval starts, to extract the named entities it seeds from, but none once traversal begins. It uses the relation label to filter hop candidates, which cuts the number of triples read at each hop. The graph holds entities only, with provenance maps linking triples and entities back to their source passages, which removes mechanism (c) while still returning original text. Traversal stays local to the neighborhood around the seed entities and never walks the whole graph, which removes mechanism (a). We did not time SHREK against HippoRAG~2, so locality here is a property of the algorithm and not a measured saving. SHREK does not break the question into sub-questions, only into entities, since each entity the NER step returns starts its own expansion.

We evaluate SHREK against two graph baselines, Dijkstra and Prize-Collecting Steiner Tree (PCST), and six reported systems, on the seven datasets of the HippoRAG~2 suite (\S\ref{sec:setup}). A local retriever gets close: SHREK has the second-highest average of the eight systems, $1.1$ points behind HippoRAG~2, and scores above our dense baseline on all four multi-hop datasets.

SHREK makes three contributions:
\begin{enumerate} 
    \item \textbf{Chain scoring without training, over an OpenIE graph.} This is what replaces the global pass and the in-loop LLM filter. SHREK scores concatenated passage chains with an off-the-shelf cross-encoder, so passages are judged by how well they complete the answer together rather than how relevant each is alone (\S\ref{sec:concat_ce}). Chain scoring is not new (\S\ref{sec:related}). SHREK differs from prior chain scorers in where the candidates come from, a relation-labeled OpenIE graph rather than hyperlinks or a fixed candidate list; in the scorer, a pretrained reranker that needs no training data; and in the cost, one cross-encoder pass per chain rather than one language model pass. Table~\ref{tab:chain_scoring} (App.~\ref{app:chain_scoring}) sets this out.
    \item \textbf{A system comparison and a failure study.} We ablate the main components of SHREK (App.~\ref{app:SHREK_tuning}, App.~\ref{app:ablation_tables}) and compare it per-question against HippoRAG~2 (\S\ref{sec:discussion}). SHREK lands $1.1$ points behind HippoRAG~2 on average, and the gap is not spread evenly: it loses 2Wiki through reach, and MuSiQue through both reach and chain scoring. Where the losses fall is more informative than the average, because the scoring limit is sharp: concatenation scoring works at 2 hops and degrades at 3 and beyond (App.~\ref{app:discrimination}).
    \item \textbf{Two graph baselines on this suite.} We give, to our knowledge, the first evaluation of PCST- and shortest-path-based passage retrieval~\citep{he2024g} on the HippoRAG~2 benchmark suite.
\end{enumerate}

\section{Related Work}
\label{sec:related}
Dense passage retrieval (DPR)~\citep{karpukhin2020dense} encodes the question and the passages into one vector space and picks the top passages by cosine similarity. To handle multi-hop questions, retrievers add a graph over the corpus, so a later-hop passage can be reached from an earlier one even when it shares few words with the question. HippoRAG~\citep{gutierrez2024hipporag} extracts OpenIE triples with an LLM. Open Information Extraction (OpenIE)~\citep{etzioni2008open} extracts $(\text{subject}, \text{relation}, \text{object})$ triples without a fixed relation schema. HippoRAG builds a graph of noun phrases from these triples and runs Personalized PageRank (PPR)~\citep{haveliwala2002topic} from seeds picked by query-side named-entity recognition (NER) and embedding matching. HippoRAG~2~\citep{gutierrez2025rag}, our main baseline, keeps the PPR walk (a) and adds mechanisms (b) and (c) from the Introduction, replacing the query-side NER with embedding matching from the query to triples. PropRAG~\citep{proprag}\footnote{PropRAG was released after the work on this paper was done.} swaps the triples for longer propositions and adds a beam search over proposition paths, but still runs PPR over the whole graph first and narrows to a subgraph after that. No other baseline in our suite has both (b) and (c), which is why it is our main point of comparison. SHREK sits in the same space as HippoRAG~2: both build a graph from OpenIE triples and match the query against the graph before traversal begins. The differences are listed in Table~\ref{tab:positioning} (App.~\ref{app:positioning}).

Other systems summarise instead of walking a graph: GraphRAG~\citep{edge2024local} answers over LLM summaries of graph communities, RAPTOR~\citep{sarthi2024raptor} builds a tree of summaries, and LightRAG~\citep{guo2024lightrag} labels its edges but matches query keywords against them rather than walking by relation type. A second group pays an LLM call per retrieval step, either interleaving chain-of-thought with retrieval~\citep{trivedi2023interleaving} or writing explicit sub-questions~\citep{press2023measuring}.

Scoring a chain of passages together, instead of each passage on its own, is a known idea in multi-hop retrieval. PathRetriever~\citep{asai2020learning} trains an RNN to pick the next paragraph given the ones already picked, and MDR~\citep{xiong2021answering} trains an encoder that rewrites the query after each hop. Beam Retrieval~\citep{zhang2024end} joins the question, the passages picked so far, and a candidate into one input, and trains an encoder with two classification heads across all hops. PromptRank~\citep{khalifa2023few} trains no retriever, and scores a whole path by how likely a language model is to produce the question after reading it. \citet{lu2023multi} build a graph of passages linked by shared entities and mine paths over it, for cross-document relation extraction rather than QA. In SHREK the candidates come from a relation-labeled OpenIE graph extracted from the passages themselves, and not from hyperlinks, TF-IDF, or a fixed candidate list, and the scorer is an off-the-shelf cross-encoder. PromptRank is training-free as well, so the difference from it is not training but cost and candidate source: PromptRank spends one language model pass on each candidate path where SHREK spends one cross-encoder pass. We do not re-run these systems here, because each is tied to its own candidate source and corpus setup, and to our knowledge none has released runs on this suite, so a direct comparison would measure the port and not the method. Table~\ref{tab:chain_scoring} (App.~\ref{app:chain_scoring}) sets out the comparison.

Our two graph-algorithm baselines reuse known methods. G-Retriever~\citep{he2024g} and OKGQA~\citep{sui2024can} both apply the PCST algorithm to pull a subgraph for the LLM, though neither was tested on the HippoRAG suite. Every graph retriever must pick seed nodes first, and embedding-based query matching is the standard choice~\citep{gutierrez2024hipporag,gutierrez2025rag,guo2024lightrag,he2024g,sui2024can}, so seed selection is common ground rather than a difference.

\section{Methodology}
\label{sec:methodology}
Given a corpus $C$ of text passages and a question $q$, retrieval returns a short ranked list of passages $P^* \subseteq C$ for a language model to read, five in all our experiments. A knowledge graph built from $C$ sits at the centre of retrieval. The graph is used only inside the retriever, so the generator only ever sees original text. Only the retriever changes across our three pipelines (\S\ref{sec:setup}); everything else is shared.

\begin{figure*}[t]
    \centering
    \includegraphics[width=0.85\textwidth]{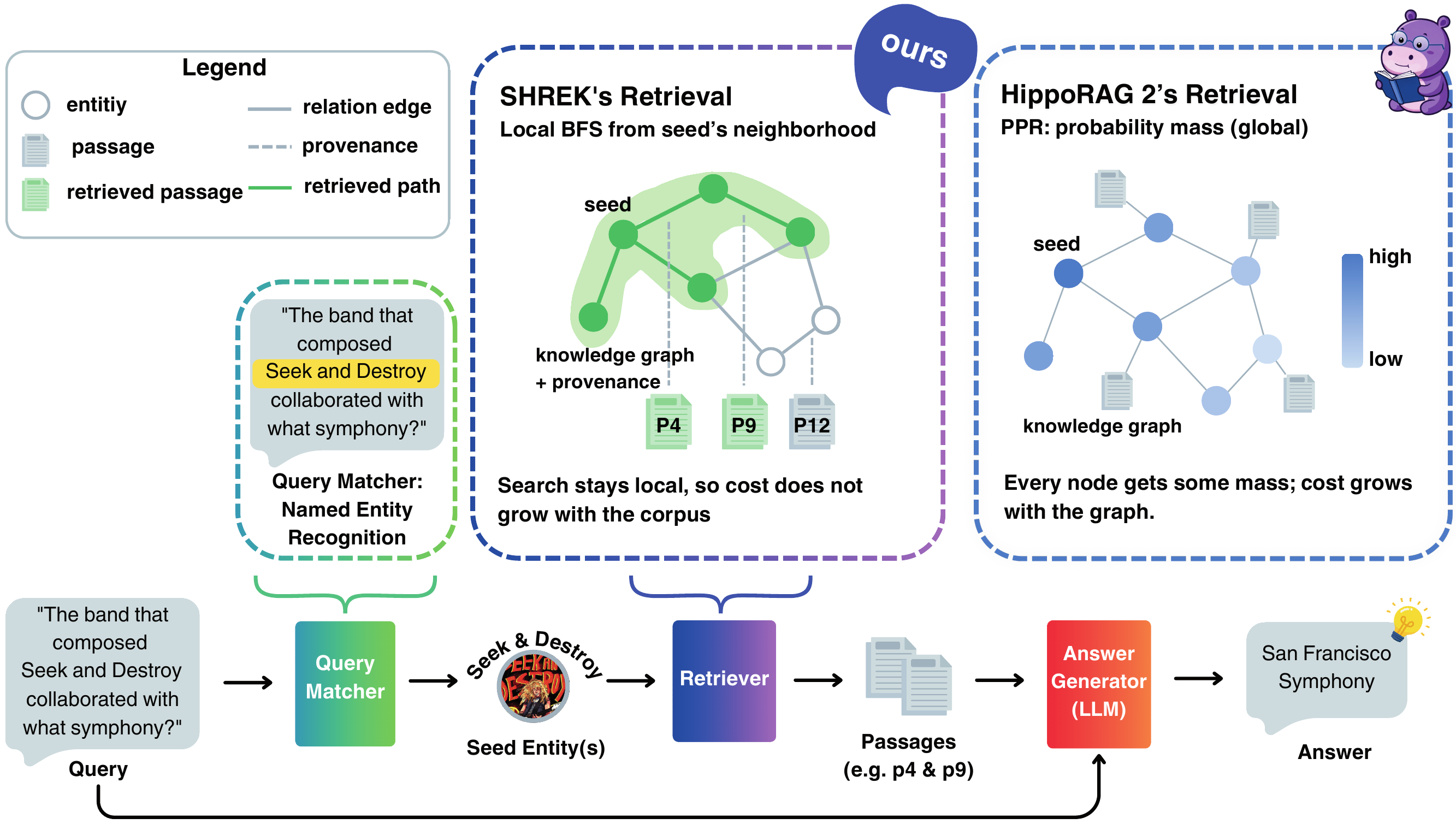}
    \caption{System overview. The pipeline runs left to right: the Query Matcher (\S\ref{sec:matchers}) turns the question into seed entities, the SHREK Retriever (\S\ref{sec:concat_ce}) expands from those seeds over the knowledge graph and returns passages through the provenance maps (\S\ref{sec:kg_construction}), and the Answer Generator (\S\ref{sec:generation}) reads only those passages. The two panels contrast the retrieval step: SHREK runs a local BFS bounded to the seed neighborhood, while HippoRAG~2, as implemented, spreads PPR mass over the whole graph.}
    \label{fig:two_component}
\end{figure*}

\paragraph{Knowledge graph.}
\label{sec:kg_construction}
We build the graph from OpenIE triples produced by HippoRAG~2's extraction code~\citep{gutierrez2025rag}, run with their settings and extractor (GPT-4o-mini) so the graph is comparable to theirs. This is one LLM pass over the corpus, before any question is asked. Let $\mathcal{G} = (\mathcal{E}, \mathcal{T})$ be the knowledge graph, with entity nodes $\mathcal{E}$ and triple edges $\mathcal{T}$. Each triple $(e_s, r, e_o)$ becomes an edge in $\mathcal{G}$, which we build as a directed multi-graph using NetworkX~\citep{hagberg2007exploring}, with entities as nodes and the relation label $r$ kept on the edge. Traversal follows edges in both directions, so an edge is usable from either endpoint. Unlike the adjacency matrix that PPR walks in HippoRAG and HippoRAG~2, this keeps the label in a form the retriever can read. The graph is entity-only, and the link back to text sits outside it in two provenance maps built during extraction: $\pi_{\mathcal{T}}$ maps each triple to the passages it came from, and $\pi_{\mathcal{E}}$ maps each entity to the passages that mention it. Both are dictionaries, so retrieval returns original passages without mixing passage chunks into the graph. We also add three edge types that OpenIE does not produce. A synonym edge joins two entities whose embeddings are near-duplicates (cosine $\geq 0.88$), such as \texttt{US} and \texttt{United States}, and App.~\ref{app:synonym} reports how much this threshold matters.

The other two edge types handle long OpenIE phrases that hold a real entity name inside them. A buried-entity edge joins the phrase to the clean name inside it, and a related-to edge links the head of the triple straight to that clean name. Synonym and buried-entity edges are not extracted facts and are excluded whenever the retriever scores or ranks triples, while a related-to edge is scored and walked like any OpenIE triple (App.~\ref{app:edge_types}). Their effect is measured in \S\ref{sec:graph_edge_ablation}.

The retriever can also merge each entity with its synonym and buried-entity neighbours before traversing, which we call alias merging. We report SHREK with it off, which works better (App.~\ref{app:SHREK_tuning}).

\paragraph{Embeddings and notation.}
Alongside the graph and the provenance maps, we precompute NV-Embed-v2 \citep{2405.17428v3} embeddings for every passage, entity, and triple, all L2-normalised so cosine similarity is a dot product. Following NV-Embed-v2's asymmetric setup, only queries carry a task instruction prefix (App.~\ref{app:prompts}). We write $\vec{q}$ for the embedding of the query $q$ and $\vec{p}$ for the embedding of a passage $p$, so the bi-encoder score is $s(\vec{q}, \vec{p}) = \vec{q} \cdot \vec{p}$.

\paragraph{Query Matcher.}
\label{sec:matchers}
The Query Matcher (Figure~\ref{fig:two_component}, left) turns the question into seed entities, and runs in two steps. The first step is named-entity recognition (NER) on the question. We prompt GPT-4o-mini once to return the entity strings the question mentions. We first tried spaCy~\citep{Honnibal_spaCy_Industrial-strength_Natural_2020} but it missed entity types these benchmarks rely on, such as song, album, and film titles. The second step is entity matching. Each extracted string is scored against every KG entity by a weighted sum of embedding cosine similarity and longest-common-subsequence overlap (App.~\ref{app:matcher}). We keep the top $5$ KG entities per string and merge them into up to $10$ seeds. This is the only LLM call SHREK makes before generation.

\begin{figure}[ht!]
    \centering
    \includegraphics[width=0.45\textwidth]{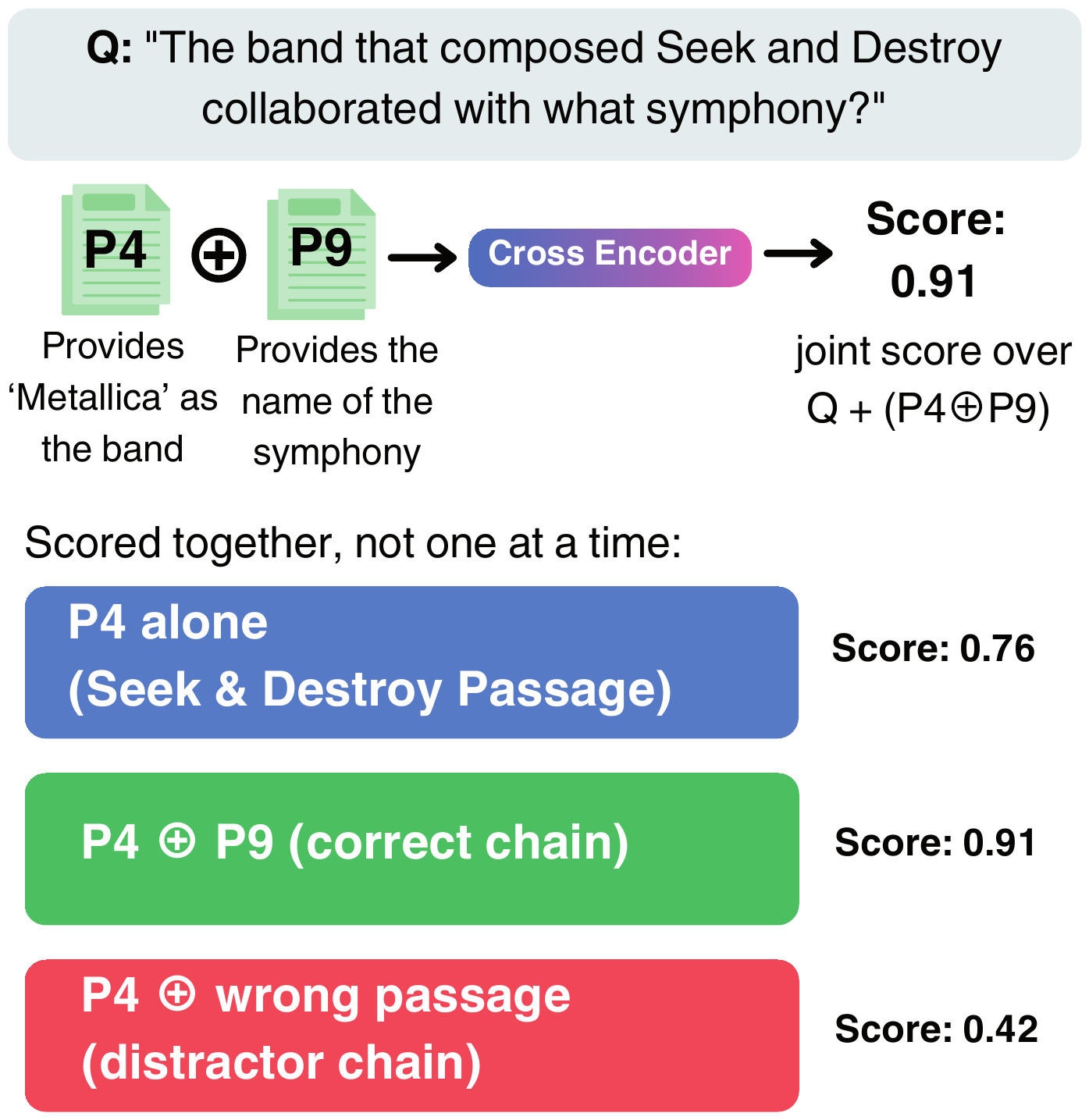}
    \caption{Chain scoring in the SHREK Retriever (\S\ref{sec:concat_ce}). The cross-encoder scores the concatenation $p_1 \oplus p_2$ against the question, not each passage alone, so the correct chain outscores both the single passage and a distractor chain even though the second passage shares few words with the question.}
    \label{fig:concat_ce}
\end{figure}

\paragraph{SHREK Retriever.}
\label{sec:concat_ce}
The SHREK Retriever (Figure~\ref{fig:two_component}, centre) expands from the seed entities over the graph and returns passages. A multi-hop question often needs two or three passages that look weakly relevant on their own. For the question ``The band that composed Seek and Destroy collaborated with what symphony?'', the answer needs the \textit{Seek and Destroy} passage, which names Metallica, and the Metallica passage, which names the San Francisco Symphony (Figure~\ref{fig:concat_ce}). The second shares almost no words with the question, so a reranker scoring one passage at a time ranks it low. The full algorithm is given in Algorithm~\ref{algo:SHREK}.

\begin{algorithm}[!ht]
\caption{SHREK retrieval\label{algo:SHREK}}
\begin{algorithmic}
\Require question $q$, seed entities $S$, graph $\mathcal{G}$
\Ensure ranked passages $P^{\ast}$

\State \textbf{Step 1: BFS expansion.}
\State $B \gets \textsc{Bfs}(\mathcal{G}, S)$, scoring each $p$ by $s(\vec{q}, \vec{p})$
\State stop once $M_{\text{anchor}}$ passages exceed $\theta_{\text{bfs}}$, else expand to max depth

\State \textbf{Step 2: Anchor selection.}
\State $B_{\text{ce}} \gets \operatorname*{top}_{N_{\text{ce}}}\, B$ \ by \ $s(\vec{q}, \vec{p})$
\State $A \gets \operatorname*{top}_{K_{\text{anchor}}}\, B_{\text{ce}}$ \ by \ $\mathrm{CE}(q, p)$

\State \textbf{Step 3: Hop expansion.}
\State $\mathit{Pairs} \gets \operatorname*{top}_{K_{\text{pair}}} \{\, p_1 \oplus p_2 : p_1 \in A,\ p_2 \in \textsc{Hop}(p_1) \,\}$ \ by \ $\mathrm{CE}(q, p_1 \oplus p_2)$
\State $\mathit{Triples} \gets \operatorname*{top}_{K_{\text{triple}}} \{\, c \oplus p_3 : c \in \mathit{Pairs},\ p_3 \in \textsc{Hop}(c) \,\}$ \ by \ $\mathrm{CE}(q, c \oplus p_3)$

\State \textbf{Step 4: Final selection.}
\State $P^{\ast} \gets \textsc{Interleave}(A, \mathit{Pairs}, \mathit{Triples})$, dedup by doc ID
\State \Return first $K_{\text{final}}$ of $P^{\ast}$
\end{algorithmic}
\end{algorithm}

$\textsc{Bfs}(\mathcal{G}, X)$ is a breadth-first search (BFS) that expands from entity set $X$ and returns the passages of the entities and triples it reaches, through $\pi_{\mathcal{E}}$ and $\pi_{\mathcal{T}}$. $\mathrm{CE}(q, \cdot)$ is a cross-encoder (\texttt{BAAI/bge-reranker-v2-m3})~\citep{chen2024m3} that reads the question and passages together, and $\oplus$ concatenates passage text. Step 2 first cuts the BFS pool to the $N_{\text{ce}}$ best passages by bi-encoder score, because the cross-encoder scores one pair at a time, which makes it the most expensive step. The $K_{\text{anchor}}$ passages kept after reranking are the \emph{anchors} that chains extend from. $\textsc{Hop}(c)$ expands from the tail entities of chain $c$'s last passage, up to a bounded depth, and returns the passages it reaches. It keeps only candidates that pass the direction filter $\theta_{\text{dir}}$, which scores the query against the embedding of the verbalized triple, and the bi-encoder prefilter $\theta_{\text{pre}}$. \textsc{Interleave} places the best anchor first, then alternates unique second-hop passages with the remaining anchors and fills leftover slots with third-hop passages, which hedges between single-hop and multi-hop questions. The first $K_{\text{final}}$ passages are returned, matching the $5$-passage budget shared across all pipelines, and all threshold and count values are in App.~\ref{app:SHREK_hparams}.

\paragraph{Answer Generator.}
\label{sec:generation}
The Answer Generator (Figure~\ref{fig:two_component}, right) takes $P^*$ and $q$ and answers with the chain-of-thought prompt from HippoRAG~2 (App.~\ref{app:test_prompt}), which asks the model to reason over the retrieved passages after ``Thought:'' and then give a short answer after ``Answer:''. Model choices are in \S\ref{sec:setup}.

\section{Experiment Setup}
\label{sec:setup}

\paragraph{Baselines.} \label{sec:dijkstra}\label{sec:pcst}
We compare against two groups. The first are reported baselines, whose numbers we take from the HippoRAG~2 paper~\citep{gutierrez2025rag}: NV-Embed-v2~(7B)~\citep{2405.17428v3} used as a dense retriever, RAPTOR~\citep{sarthi2024raptor}, GraphRAG~\citep{edge2024local}, LightRAG~\citep{guo2024lightrag}, HippoRAG~\citep{gutierrez2024hipporag}, and HippoRAG~2 itself. That paper uses the same benchmark suite, generator, and embedding model as our runs. The second are baselines we implement and run ourselves under our exact configuration: the same NV-Embed-v2 dense retriever as a no-graph reference point, and two graph-algorithm pipelines built over the same knowledge graph and embeddings as SHREK (\S\ref{sec:related}). \emph{Dijkstra} treats retrieval as shortest-path search between the endpoints of query-matched triples, and \emph{PCST} re-implements PCST retrieval~\citep{sui2024can} over a two-hop expansion from the seeds. Both deduplicate, rank by query-passage similarity, and return the top passages. App.~\ref{app:baseline_details} gives the full description of each.

\paragraph{Datasets.}
We evaluate on the seven test datasets from the HippoRAG~2 suite, listed in Table~\ref{tab:dataset_stats}. Two are simple question answering (QA), where the answer usually sits in a single passage: PopQA~\citep{mallen2023not} (entity questions) and Natural Questions, or NQ~\citep{kwiatkowski2019natural} (open-domain). Three are multi-hop, where the answer needs two or more passages: HotpotQA \citep{yang2018hotpotqa}, 2WikiMultihopQA, or 2Wiki~\citep{ho2020constructing}, and MuSiQue~\citep{trivedi2022musique}. LV-Eval~\citep{yuan2024lv} is long-context multi-hop, and NarrativeQA~\citep{kovcisky2018narrativeqa} tests reading over long single-document narratives. 

\paragraph{Validation split.} \label{sec:val_construction}
We build a $500$-question validation set for each of PopQA, HotpotQA, and MuSiQue. These are disjoint from the HippoRAG test splits: we sample from each official training set with a fixed seed (\texttt{random\_state=42}) and drop any question whose ID appears in the HippoRAG test pool. Sampling is uniform, so the MuSiQue validation sample keeps the native 2-, 3-, and 4-hop mix of its training split. All tuning runs on these splits, and every number in \S\ref{sec:results} comes from the held-out test splits. We apply one configuration to all seven test sets rather than tuning per dataset, and the Limitations give our reason and the evidence.

\paragraph{Metrics.}
Our primary metric is Token~F1, the token overlap between the predicted and gold answer, both lowercased and stripped of punctuation. When a question has several gold answers (as in PopQA aliases), we take the maximum F1 across them and then average across questions. We also report BERTScore~\citep{bert-score} in App.~\ref{app:bertscore}. For retrieval, we use Recall@5, the fraction of gold supporting passages that appear in the top $5$ retrieved passages. A passage counts as gold only if the dataset marks it as a supporting passage. Some ablations report Recall@10 instead, which is more sensitive to a single component change. 

\paragraph{Models.}
All pipelines share the same embedding model, cross-encoder, and generator, so any difference in results comes from the retrieval algorithm and not the surrounding parts. Embeddings for passages, entities, and triples use NV-Embed-v2~\citep{2405.17428v3}. The cross-encoder inside SHREK is \texttt{BAAI/bge-reranker-v2-m3}~\citep{chen2024m3}. GPT-4o-mini is used everywhere an LLM is needed: the OpenIE extraction, the NER and triple-filter calls, and generation for the main test-set results. This matches the GPT-4o-mini setting HippoRAG 2 reports. For all validation-set tuning we use Llama-3.3-70B-Instruct with a simpler answer prompt (App.~\ref{app:val_prompt}). The test runs use GPT-4o-mini with HippoRAG~2's chain-of-thought prompt (App.~\ref{app:test_prompt}). So the generator and the prompt both differ between tuning and test, and we did not re-tune after the switch (see Limitations). Both run at temperature $0$. All retrieval runs use one NVIDIA L40S 48GB GPU, and generation calls are served through OpenRouter~\citep{openrouter}.

\section{Results}
\label{sec:results}

Table~\ref{tab:all_benchmarks} reports test-set Token F1. Our three pipelines each use the best configuration found on the validation split (App.~\ref{app:SHREK_tuning}).

\begin{table*}[t]
\centering
\resizebox{\textwidth}{!}{%
\begin{tabular}{lcccccccc}
\toprule
& \multicolumn{2}{c}{Simple QA}
& \multicolumn{4}{c}{Multi-Hop QA}
& \multicolumn{1}{c}{\shortstack{Discourse\\Understanding}} & \\
\cmidrule(lr){2-3} \cmidrule(lr){4-7} \cmidrule(lr){8-8}
Pipeline & NQ & PopQA & MuSiQue & 2Wiki & HotpotQA & LV-Eval 
& NarrativeQA & Avg \\
\midrule
\multicolumn{9}{c}{\bf Numbers taken from \cite{gutierrez2025rag}} \\ \midrule
NV-Embed-v2 (7B) \cite{2405.17428v3}
  & 59.9 & 55.8 & 46.0 & 60.8 & 71.0 & 10.0 & 24.2 & 46.8 \\
RAPTOR \cite{sarthi2024raptor}               
  & 54.5 & 55.1 & 39.2 & 48.4 & 64.7 &  9.2 & 21.8 & 41.8 \\
GraphRAG \cite{edge2024local}             
  & 55.5 & 51.3 & 42.0 & 61.0 & 67.6 & 11.0 & 20.9 & 44.2 \\
LightRAG \cite{guo2024lightrag}             
  & 15.4 & 14.8 &  9.3 & 12.1 & 20.2 &  5.0 &  9.0 & 12.3 \\
HippoRAG \cite{gutierrez2024hipporag}             
  & 52.2 & 56.2 & 35.9 & 67.3 & 60.0 &  7.6 & 16.1 & 42.2 \\
HippoRAG 2 \cite{gutierrez2025rag}           
  & 60.0 & 55.7 & 49.3 & 69.7 & 71.1 & 14.0 & 25.2 & 49.3 \\
\midrule
\multicolumn{9}{c}{\bf Our Results} \\ \midrule
NV-Embed-v2 (7B) (Ours)     
 & \textbf{61.7} & 61.8 & 44.2 & 57.2 & 72.2 & 11.7 & \textbf{26.2} & 47.8 \\
Dijkstra pathfinding 
  & 58.3 & 58.6 & 35.6 & 41.9 & 57.6 & 8.4 & 22.0 & 40.3 \\
PCST                 
  & 60.8 & 62.1 & 44.6 & 55.4 & 70.6 & 9.2 & 26.1 & 47.0 \\
\textbf{SHREK}     
  & 59.4 & \underline{\textbf{62.3}} & \textbf{45.8} & \textbf{60.0}$^\dagger$ & \underline{\textbf{73.0}}$^\dagger$ 
  & \textbf{12.2} & 24.5 & \textbf{48.2} \\
\bottomrule
\end{tabular}
} 
\caption{Token F1 on the seven test sets of the HippoRAG~2 suite. All systems use NV-Embed-v2~\cite{2405.17428v3} embeddings, GPT-4o-mini for generation, the same chain-of-thought prompt (App.~\ref{app:prompts}), and the top $5$ passages as context. Upper rows are as reported in~\citep{gutierrez2025rag}, taking the GPT-4o-mini setting so the extractor and generator match ours; lower rows are our runs. In the lower block, bold marks the best per dataset, underline marks where SHREK scores higher than the reported HippoRAG~2 value, and $^\dagger$ marks where SHREK beats PCST significantly (paired bootstrap, Holm-corrected, $p < 0.05$; App.~\ref{app:significance}). The upper rows come with no per-question outputs, so we cannot test them; App.~\ref{app:agreement} tests SHREK against our own rerun of HippoRAG~2. Avg is the unweighted mean over the seven values for every row, so it differs from the question-weighted average in~\citep{gutierrez2025rag}.}
\label{tab:all_benchmarks}
\end{table*}

Table~\ref{tab:recall} reports Recall@5 on the five datasets for which HippoRAG~2 reports recall.

\begin{table*}[t]
\centering
\small
\begin{tabular}{lccccc}
\toprule
Pipeline & NQ & PopQA & MuSiQue & 2Wiki & HotpotQA \\
\midrule
RAPTOR \cite{sarthi2024raptor}     
  & 69.4 & 48.1 & 61.0 & 66.0 & 90.2 \\
HippoRAG \cite{gutierrez2024hipporag}   
  & 45.1 & 52.2 & 52.4 & 87.0 & 78.5 \\
HippoRAG 2 \cite{gutierrez2025rag} 
  & \textbf{76.4} & 52.2 & \textbf{74.2} & \textbf{90.2} 
  & \textbf{95.7} \\
\midrule
NV-Embed-v2 (7B) (Ours)
  & 75.5 & 50.4 & 67.4 & 75.1 & 92.9 \\
Dijkstra pathfinding   
  & 50.7 & 39.8 & 42.4 & 53.5 & 62.4 \\
PCST                   
  & 75.1$^\dagger$ & 50.3 & 67.2$^\dagger$ & 73.8 & 91.5 \\
\textbf{SHREK}                
  & 63.2 & \textbf{53.9}$^\dagger$ & 64.4 & 80.7$^\dagger$ & 92.8 \\
\bottomrule
\end{tabular}
\caption{Recall@5 against gold supporting passages on the five datasets for which HippoRAG~2 reports recall. A passage counts as gold only if the dataset marks it as supporting. All systems retrieve the top $5$ passages. Upper rows are as reported in~\citep{gutierrez2025rag}; lower rows are our runs. Bold marks the best per column, and $^\dagger$ marks the higher of SHREK and PCST where the bootstrap separates them (Holm-corrected, $p < 0.05$; App.~\ref{app:significance}). The upper rows carry no markers, for the reason given in Table~\ref{tab:all_benchmarks}.}
\label{tab:recall}
\end{table*}

\paragraph{Answer quality.} SHREK has the second-highest average of the eight systems in Table~\ref{tab:all_benchmarks}, $1.1$ points behind HippoRAG~2, and is the strongest of our three pipelines. It beats HippoRAG~2 on PopQA ($+6.6$), though PopQA is not a retrieval win, since recall is close there ($53.9$ vs $52.2$) and \S\ref{sec:discussion} points to answer length. The $1.9$ point lead on HotpotQA is not significant (\S\ref{sec:discussion}). The gap is larger on the harder multi-hop sets, MuSiQue ($-3.5$) and 2Wiki ($-9.7$). PCST is competitive on simple QA but trails SHREK significantly on 2Wiki ($4.6$ points) and HotpotQA ($2.5$), while Dijkstra is weakest throughout, trailing by $15$ to $18$ points on the same two sets, since its single-path output cannot cover the multiple supporting passages these datasets need. SHREK matches or beats PCST across the multi-hop sets while keeping retrieval local.

\paragraph{Retrieval recall.} HippoRAG~2 has the highest Recall@5 on four of five datasets (Table~\ref{tab:recall}). SHREK beats it only on PopQA and is close on HotpotQA. The three losses are similar in size, $13.2$ recall points on NQ, $9.8$ on MuSiQue and $9.5$ on 2Wiki, but they cost very different amounts of answer quality: $0.6$ Token F1 on NQ, $3.5$ on MuSiQue and $9.7$ on 2Wiki. The size of the recall gap therefore does not predict the size of the F1 gap, and what matters is which passages are missing. On NQ the small F1 gap suggests the generator often answers without the gold passage. On 2Wiki the missing passages are the second hop of bridge-comparison and inference questions, which reward connecting entities through indirect paths that a local neighborhood does not reach, so the scorer never sees them. On MuSiQue the failure analysis in \S\ref{sec:discussion} finds losses at both stages, reach and chain scoring. 

Hereafter, we report ablations on the three validation splits, using the same configuration as above except with Llama-3.3-70B-Instruct in place of the test-set LLM. Validation tables report Token F1 on a $0$ to $1$ scale, while the test-set tables above use a $0$ to $100$ scale.

\paragraph{Graph enrichment.} \label{sec:graph_edge_ablation} We compare a base graph (OpenIE triples only) against an enriched graph on the three validation sets (Table~\ref{tab:graph_enrichment}). For SHREK the enriched graph improves Token F1 by $+1.1$ on PopQA, $+1.3$ on MuSiQue, and $+0.9$ on HotpotQA, a small but consistent gain that holds with alias merging off. The likely reason is the related-to edges, which are walked like ordinary edges and raise the walkable edge count by $8$ to $9\%$.

\paragraph{The direction filter.} \label{sec:direction_ablation} The direction filter is the one place the retriever reads the relation label. Lowering $\theta_{\text{dir}}$ until every candidate passes moves Recall@5 only from $0.762$ to $0.758$ while mean retrieval time rises by $40\%$ (App.~\ref{app:direction}), so the filter saves time rather than finding better passages.

\paragraph{The cross-encoder is the mechanism.} The cross-encoder is the most expensive part of SHREK, since it needs one forward pass per pair while passage embeddings are precomputed, so we test whether the cheaper bi-encoder can replace it by scoring the concatenated passage text against the question with cosine similarity instead (Table~\ref{tab:ce_ablation}). It fails almost completely: on PopQA Token F1 drops from $0.771$ to $0.030$, close to zero. The bi-encoder encodes the question and passages independently, so it only sees a longer block of text and scores surface similarity. The cross-encoder reads them together, which is what lets a chain outscore its parts.

\paragraph{Generator comparison.} Holding retrieval fixed (Table~\ref{tab:generator}), GPT-4o-mini scores higher on single-hop PopQA ($+1.1$) while Llama-3.3-70B-Instruct scores higher on MuSiQue ($+7.8$) and HotpotQA ($+7.3$).

\section{Failure Analysis}
\label{sec:discussion}
We use MuSiQue here because it is the one validation set with explicit hop labels and gold supporting passages. We describe four ways the pipeline loses a question. They are not a partition of one set: (a) and (b) use per-stage recall over all 500 validation questions, while (c) and (d) use the 42 questions where the LLM refused to answer. 

\begin{enumerate}[label=(\alph*)]
\item \textbf{Not reachable.} At least one gold passage never appears in the BFS pool. This could come from seeds that start BFS in the wrong neighborhood, gold passage entities missing from the graph, or connecting triples OpenIE never extracted, and the stage recall data cannot separate these. The BFS pool holds all gold passages for 68.4\% of questions, so 31.6\% lose at least one before any scoring happens.
\item \textbf{Lost during scoring and selection.} The gold passage is in the BFS pool but does not survive into the final set. The bi-encoder top-30 cutoff costs 5.3\% recall, cross-encoder anchor selection costs 11\%, and the 5-passage budget pushes others out. Hop expansion recovers 5.6\% through tail-entity expansion.
\item \textbf{Partial retrieval.} The pipeline finds the first supporting passage but not the second, in 32 of the 42 refusals (76.2\%). The second passage shares few terms with the question, so even inside the BFS pool it is outranked by topically similar distractors.
\item \textbf{Generator failure.} All gold passages are in the final set but the LLM still answers wrong. Of the 42 refusals, 5 (11.9\%) fall here. These are not retrieval failures. The remaining 5 refusals fall into neither group.
\end{enumerate}

Reachability is the largest loss we measure over all 500 questions, at $31.6\%$, and it happens before any scoring, so no later step can recover from it. Partial retrieval is the most common pattern among the 42 refusals, but those are a small subset of SHREK's errors, so we do not read it as the dominant failure mode. It does match App.~\ref{app:discrimination}: the cross-encoder finds the first supporting passage but struggles with the second, and worse at higher hops. Seed quality contributes: the query matcher extracts only 1.75 named entities on average, and 25\% of questions do not have the correct named entity among the seeds.

\textbf{Comparison with HippoRAG~2}: We now run HippoRAG~2\footnote{The code/data artifacts from the paper do not include per-question outputs. Therefore, we use their provided code.} and SHREK on six of the seven test sets and sort each question into both correct, both wrong, HippoRAG~2 only, or SHREK only (App.~\ref{app:agreement}, Table~\ref{tab:buckets}), counting a question correct at Token F1 $\geq 0.5$. Both systems use the same generator, so retrieval drives almost all of these disagreements. An exact McNemar test on the two ``only'' groups, Holm-corrected across the six datasets, separates them on PopQA, where SHREK is ahead, and on 2Wiki, where it is behind. On MuSiQue, the test does not separate them after correction ($p_{\text{Holm}} = 0.068$), and the other three sets are not separable. HippoRAG~2's largest edge is 2Wiki, which it uniquely solves 15.7\% of the time against SHREK's 4.6\%, and the gap is widest on bridge-comparison (26.8\% unique) and inference questions (20.4\%). It also leads on NQ, likely because its casual Google queries carry few named entities for our NER-based seeding, which we did not measure. SHREK's largest edge is PopQA, uniquely solving 9.2\% against 3.3\%; recall is similar there, and the difference is answer length, 2.1 words on average against 4.2 for HippoRAG~2's wrong answers, which suggests noisier context makes the generator hedge. SHREK also leads on HotpotQA comparison questions (12.7\% unique vs 3.7\%), where interleaved selection naturally places both entities' passages in the top 5. Finally, SHREK returns zero passages when the seed matcher finds nothing in the graph, on 22 questions in NQ, 13 in MuSiQue, and 53 in NarrativeQA. HippoRAG~2 always returns 5 through its dense fallback.

\section{Conclusion \& Future Work}

This work asked how far a graph-based passage retriever gets when it stays local to the seed entities, without a dense retrieval fallback and without explicit question decomposition. A local retriever gets close: SHREK has the second-highest average of the eight systems compared, $1.1$ points behind HippoRAG~2, and scores above our dense retrieval baseline on all four multi-hop datasets. The two limits are reach and concatenation scoring, which prefers the gold chain at $2$ hops but not at $3$ and $4$, so future work may use question decomposition or a task-specific scoring model.

\section*{Limitations}

\paragraph{A 2-hop ceiling.} The cross-encoder discrimination gap is positive at 2 hops and negative at 3 and 4 hops (App.~\ref{app:discrimination}), so chain scoring stops being reliable past two hops. The 2Wiki gap has a different cause, reachability rather than scoring, since its questions are 2-hop but often need entities that a local neighborhood does not reach.

\paragraph{The generator and the prompt differ between tuning and test.} We tuned SHREK on the validation splits with Llama-3.3-70B-Instruct and a simple answer prompt (App.~\ref{app:val_prompt}), and report the main test results with GPT-4o-mini and HippoRAG~2's chain-of-thought prompt (App.~\ref{app:test_prompt}), to match HippoRAG~2's reporting setup. We did not re-tune after the switch. Table~\ref{tab:generator} shows the two generators are not interchangeable on identical retrieval: GPT-4o-mini is better on single-hop PopQA ($+1.1$) but worse on MuSiQue ($-7.8$) and HotpotQA ($-7.3$), so the generator we report with is the weaker of the two on two of the three datasets we can tune on. This affects our Token F1 results in two ways. The multi-hop test scores likely understate what SHREK's retrieval provides, and the configuration we report was chosen under a different generator and prompt, so another configuration might suit GPT-4o-mini better. Recall@5 does not depend on the generator, so Table~\ref{tab:recall} is unaffected. Re-tuning under the test generator is the obvious check and we did not run it.

\paragraph{One configuration across all seven datasets.} We only have validation splits for PopQA, MuSiQue, and HotpotQA, and apply those tuning decisions unchanged to the other four test sets (NQ, 2Wiki, NarrativeQA, LV-Eval). This is a deliberate trade. Tuning per dataset would need a validation split sampled from each remaining training set, and it risks reporting a configuration fitted to each test distribution rather than a method that transfers. HippoRAG~2 reports a single configuration across the same suite, so holding to one keeps the comparison on the same footing. Where we can measure the cost, it is small. The right measure is not the spread across configurations but the gap between the configuration we report and the best cell for each dataset in the tuning grid (App.~\ref{app:SHREK_tuning}). On MuSiQue ($0.520$) and HotpotQA ($0.755$) the configuration we report is already the best cell, and on PopQA it sits $0.3$ Token F1 points below the best ($0.771$ against $0.774$). So on the three datasets we can tune, the gain from per-dataset tuning is $0.3$ points on one set and nothing on the other two. The grid varies one component at a time and does not cover every combination, so this is a lower bound on what an exhaustive search might find. More to the point, the two datasets we lose most on fail for reasons tuning does not touch. NQ uses casual Google queries with few named entities, so the seed matcher starts in the wrong neighborhood or returns nothing at all. 2Wiki has bridge-comparison and inference question types that none of our validation sets contain, and needs entities a local neighborhood never reaches. Neither is a threshold problem. A per-dataset configuration would therefore help a little where the pipeline already works, and would close neither gap. We did not run that comparison, and it remains open.

\paragraph{We did not measure efficiency.} We call SHREK local because of what the algorithm visits, not because of a measured saving. The running time of PPR depends on the implementation, on how many seed nodes carry non-zero mass, and on whether the walk is run to convergence, so we cannot claim in general that a global walk costs more than a local one. SHREK is not uniformly local either: its seed matcher scores each extracted string against every entity in the graph, which is a global pass before traversal starts. The one claim we do make about the LLM is placement and not count. HippoRAG~2 also makes one call per question, but it sits inside retrieval and reads candidate triples pulled from the graph, while SHREK's happens before retrieval and reads only the question. A timed comparison against HippoRAG~2 under matched hardware and implementations is left to future work, and until it is run the locality argument stands as a property of the algorithm and not an efficiency result.

\paragraph{Significance testing is partial.} We test SHREK against PCST with a paired bootstrap on Token F1 and Recall@5 (App.~\ref{app:significance}), and against our own rerun of HippoRAG~2 with an exact McNemar test at the question level (Table~\ref{tab:buckets}). Everything else in Tables~\ref{tab:all_benchmarks} and~\ref{tab:recall} is a gap between means and not a tested difference. We cannot test the reported HippoRAG~2 rows at all, because \citet{gutierrez2025rag} release no per-question outputs, and we did not test our dense baseline. The McNemar test has its own limits: it counts a question correct at Token F1 $\geq 0.5$, so it does not test mean Token F1, and it covers neither Recall@5 nor LV-Eval, which we did not rerun. LV-Eval ($124$ questions) and NarrativeQA ($293$) are small, so a result the test cannot separate there means it cannot tell the two systems apart, not that they perform the same.

\paragraph{Relation labels help pruning, not accuracy.} We keep relation labels on every edge, and the direction filter (\S\ref{sec:concat_ce}) is the one place the retriever reads them. Turning that filter off costs $2$ Recall@10 points and $40\%$ more retrieval time (\S\ref{sec:direction_ablation}), but the text the filter scores also carries the subject and the object, so our claim about relation labels is about pruning and not about answer quality.

\paragraph{No dense fallback.} SHREK has no dense retrieval fallback, so when the seed matcher finds nothing the retrieval step returns zero passages, on 22 questions in NQ, 13 in MuSiQue, and 53 in NarrativeQA. We left the fallback out so the contribution of graph-based retrieval stays clean, but it is an obvious improvement prior to a potential deployment.

\paragraph{Extraction quality bounds every graph method here.} Our graph comes entirely from the OpenIE triples produced by HippoRAG~2's extraction code, which we ran with GPT-4o-mini under their settings. Every graph-based method in our comparison uses the same triples, so the comparison is fair, but the absolute performance of every graph-based method is bounded by extraction quality. It also means that when we say SHREK makes no LLM call inside the retrieval loop, we are talking about the per-query cost only. The system as a whole still pays for an LLM pass over the corpus first. That cost is paid once and is shared by every graph-based method here, but it should be counted. Better extraction, such as relation-aware extractors or typed entity schemas, would likely lift all graph-based methods.

\paragraph{PopQA validation and test differ.} SHREK scores higher on the PopQA validation split than on the PopQA test split. Some of this is the generator change, but the likely remaining cause is that the test set has more long-tail entities with weaker graph coverage.

\section*{Ethical Considerations}
Our work reuses public QA benchmarks and OpenIE triples produced with HippoRAG~2's extraction code. We do not collect or label any data ourselves, so the work raises no annotation-labour or privacy concerns beyond those already present in the source datasets, which are built from public sources such as Wikipedia. As with any RAG system, the generator can still produce a wrong answer when retrieval fails, as our failure analysis shows, so outputs should not be treated as authoritative without checking the retrieved passages.

\section*{Acknowledgment}
This paper is the outcome of a Taste-of-Research scholarship awarded to Dylan Uno Syahfaril by the Faculty of Engineering at UNSW Sydney. The authors would like to thank Pantelis Elinas for his guidance and contribution throughout this project. This research includes computations using the computational cluster Katana supported by Research Technology Services at UNSW Sydney.
\bibliography{custom}

\appendix

\section{Positioning}
\subsection{Comparison of SHREK with previous approaches}
\label{app:positioning}
Table~\ref{tab:positioning} compares SHREK with closely related approaches.
\begin{table*}[t]
\centering
\small
\setlength{\tabcolsep}{5pt}
\begin{tabular}{lccccc}
\toprule
System & Retrieval & \shortstack{Reads relation\\labels} & Scope & \shortstack{LLM in\\retrieval loop} & \shortstack{Dense\\fallback} \\
\midrule
Dense (NV-Embed-v2)~\citep{2405.17428v3} & Dense  & \ding{55}          & Global & \ding{55} & n/a       \\
HippoRAG~\citep{gutierrez2024hipporag}   & Graph  & \ding{55}          & Global & \ding{55} & \cmark    \\
LightRAG~\citep{guo2024lightrag}         & Graph  & \ding{55}$^\dagger$ & Global   & \cmark    & \ding{55} \\
HippoRAG~2~\citep{gutierrez2025rag}      & Graph  & \ding{55}          & Global & \cmark    & \cmark    \\
PropRAG~\citep{proprag}                  & Graph  & \ding{55}          & Global & \ding{55} & \ding{55} \\
\midrule
SHREK (ours)                          & Graph  & \cmark             & Local  & \ding{55} & \ding{55} \\
\bottomrule
\end{tabular}
\caption{Positioning of SHREK against representative retrievers. \emph{Reads relation labels} means the retriever traverses the graph using edge relation labels, not just connectivity. Both HippoRAG variants match the query against triple embeddings before traversal begins, so the label reaches them at seed selection, but not during the walk itself. \emph{Scope} is whether traversal stays in the neighbourhood around the seed entities (Local) or covers the whole graph before passages are picked (Global). It describes traversal only. SHREK's seed matching still scores each extracted string against every entity in the graph. \emph{LLM in retrieval loop} counts only calls made after traversal starts. It excludes any query-side NER or triple-filter call made before that point. $^\dagger$LightRAG stores relation labels on edges but matches against them rather than walking by relation type.}
\label{tab:positioning}
\end{table*}

\subsection{Comparison with chain-scoring retrievers}
\label{app:chain_scoring}

Table~\ref{tab:chain_scoring} compares SHREK with the chain-scoring retrievers discussed in \S\ref{sec:related}, along four axes: where candidate chains come from, what scores them, whether training data is needed, and what one chain costs to score.

\begin{table*}[t]
\centering
\small
\setlength{\tabcolsep}{5pt}
\begin{tabular}{@{}
  >{\raggedright\arraybackslash}p{\dimexpr 0.22\textwidth - 8pt\relax}
  >{\raggedright\arraybackslash}p{\dimexpr 0.24\textwidth - 8pt\relax}
  >{\raggedright\arraybackslash}p{\dimexpr 0.20\textwidth - 8pt\relax}
  >{\centering\arraybackslash}p{\dimexpr 0.10\textwidth - 8pt\relax}
  >{\raggedright\arraybackslash}p{\dimexpr 0.24\textwidth - 8pt\relax}
@{}}
\toprule
System & Candidate source & Scorer & \shortstack{Training\\data} & Cost per chain \\
\midrule
PathRetriever~\citep{asai2020learning} & Wikipedia hyperlink graph & Trained RNN & \cmark & One RNN step \\
MDR~\citep{xiong2021answering} & Dense index, requeried per hop & Trained bi-encoder & \cmark & One encoder pass per hop \\
Beam Retrieval~\citep{zhang2024end} & Fixed per-dataset candidate list & Trained encoder, two heads & \cmark & One encoder pass \\
PromptRank~\citep{khalifa2023few} & Hyperlink paths & LM likelihood of the question & \ding{55} & One language model pass \\
\midrule
SHREK (ours) & Relation-labeled OpenIE graph & Pretrained cross-encoder & \ding{55} & One cross-encoder pass \\
\bottomrule
\end{tabular}
\caption{SHREK against prior chain-scoring retrievers. \emph{Training data} is whether the scorer must be trained on question-passage supervision before use. PromptRank is also training-free, so the difference there is the candidate source and the cost of scoring one chain. PathRetriever also walks a graph over the corpus, but its edges are Wikipedia hyperlinks, whereas SHREK's are OpenIE triples extracted from the passage text, so the graph exists for any corpus and carries a relation label on every edge. We report where chain scoring stops working for SHREK in App.~\ref{app:discrimination}; to our knowledge the other systems do not report an equivalent measurement.}
\label{tab:chain_scoring}
\end{table*}

\section{Implementation Details}

\subsection{Buried-entity and related-to edges}
\label{app:edge_types}

OpenIE sometimes returns a long phrase where an entity name should be, such as a tail reading ``goal bringing comparison to diego maradona\ldots'' when \texttt{diego maradona} has no node of its own. A buried-entity edge joins the long tail to the clean name inside it. Next to each buried-entity edge we add one related-to edge, which links the head of the triple straight to that clean name, so the next hop can be reached without going through the messy tail. Synonym and buried-entity edges are not extracted facts, so both are excluded whenever the retriever scores or ranks triples. A related-to edge is an ordinary edge, and is scored, ranked and walked like any OpenIE triple.

\subsection{SHREK hyperparameters}
\label{app:SHREK_hparams}

Table~\ref{tab:SHREK_hparams} lists the fixed threshold and count settings used by SHREK in the main results (\S\ref{sec:results}). The thresholds were fixed from earlier diagnostic work; the count settings were selected on the validation split (App.~\ref{app:SHREK_tuning}).

\begin{table*}[t]
\centering
\small
\begin{tabular}{lll}
\toprule
Symbol & Value & Role \\
\midrule
$\theta_{\text{bfs}}$   & 0.25 & BFS bi-encoder threshold for early stop \\
$M_{\text{anchor}}$     & 5    & Passages above $\theta_{\text{bfs}}$ to stop BFS \\
Max BFS depth           & 3    & Maximum BFS depth if early stop not hit \\
$N_{\text{ce}}$                & 30   & BFS passages reranked by cross-encoder \\
$K_{\text{anchor}}$     & 10   & Anchors kept after reranking \\
$\theta_{\text{dir}}$   & 0.15 & Direction filter on triple fact embedding \\
$\theta_{\text{pre}}$   & 0.15 & Bi-encoder prefilter on candidate passage \\
$K_{\text{pair}}$       & 5    & Pairs kept per hop expansion \\
$K_{\text{triple}}$     & 3    & Top pairs carried to hop-3 scoring \\
$K_{\text{final}}$      & 5    & Passages returned as $P^*$ \\
\bottomrule
\end{tabular}
\caption{Fixed hyperparameters for SHREK.}
\label{tab:SHREK_hparams}
\end{table*}

\subsection{Query matcher scoring}
\label{app:matcher}

This section gives the full NER + cosine + LCS scoring used in the query matcher (\S\ref{sec:matchers}). Given an NER string $n$ with embedding $\vec{n}$, each KG entity $e$ is scored:
\begin{equation*}
\begin{aligned}
r(n, e) &= 0.5\, s(\vec{n}, \vec{e}) + 0.5\, \mathrm{LCS}(n, e) \\
  &\quad + b(n, e) \\
\text{score}(n, e) &= \min\bigl(1.0,\ r(n, e)\bigr)
\end{aligned}
\end{equation*}

\noindent where:
\begin{itemize}[leftmargin=*,itemsep=2pt]
\item $s(\vec{n}, \vec{e}) = \vec{n} \cdot \vec{e}$ is the 
  cosine similarity between the two embeddings (both are 
  L2-normalised at load time, so the dot product equals 
  cosine).
\item $\mathrm{LCS}(n, e)$ is the character-level longest 
  common subsequence length between the two surface strings, 
  normalised by the length of the longer string.
\item $b(n, e) = 0.3$ if $n$ is a substring of $e$, else $0$.
\end{itemize}

The containment bonus exists to fix a failure mode of LCS. A short NER span like ``nasir zaidi'' sitting inside a long KG name like ``syed mohammad nasir zaidi'' gets a low LCS ratio (the denominator is the length of the long name), so cosine alone would have to carry the match. The bonus restores priority in that case. The final minimum with $1.0$ is just a safety clip so the score never goes above $1$.

\subsection{Baseline implementations}
\label{app:baseline_details}

Both graph baselines read the same graph, provenance maps, and embeddings as SHREK (\S\ref{sec:methodology}).

\paragraph{Dijkstra.} Its query matcher scores $\vec{q}$ against every triple embedding, passes the top candidates to GPT-4o-mini to drop irrelevant ones, and takes the head and tail entities of the survivors as seeds. For each (head, tail) pair it runs weighted Dijkstra between the endpoints, with edge weight $1 - s(\vec{q}, \vec{p}^{\,k})$ so the search prefers edges whose passages look like the query, then returns the passages along the paths. The prompt for the triple filter is in App.~\ref{app:triple_prompt}.

\paragraph{PCST.} It expands two hops from the seeds, assigns node and edge prizes from query similarity, runs the \texttt{pcst\_fast} solver (with G-Retriever's virtual-node trick so prized edges are still selected), and maps the returned tree to passages.

\section{Datasets}
\label{app:dataset_stats}

Table~\ref{tab:dataset_stats} gives the size and hop profile of every test and validation split.
\begin{table*}[t]
\centering
\small
\begin{tabular}{llrrrrrrl}
\toprule
Dataset & Split & \#Q & \#Psg & 1h & 2h & 3h & 4h & Category \\
\midrule
PopQA              & test & 1{,}000 & 8{,}676  & 1{,}000 & --    & --   & --  & single-hop (entity) \\
NQ          & test & 1{,}000 & 9{,}633  & 1{,}000 & --    & --   & --  & single-hop (open-domain) \\
HotpotQA           & test & 1{,}000 & 9{,}811  & --      & 1{,}000 & --   & --  & multi-hop (bridge/comp.) \\
2WikiMultihopQA    & test & 1{,}000 & 6{,}119  & --      & 1{,}000 & --   & --  & multi-hop (comp./bridge/infer.) \\
MuSiQue            & test & 1{,}000 & 11{,}656 & --      & 518   & 316  & 166 & multi-hop (2/3/4-hop) \\
LV-Eval            & test & 124     & 22{,}849 & --      & 124   & --   & --  & long-context \\
NarrativeQA        & test & 293     & 4{,}111  & \multicolumn{4}{c}{n/a} & narrative / single-doc \\
\midrule
PopQA-val          & val  & 500     & 933      & 500     & --    & --   & --  & single-hop (entity) \\
HotpotQA-val       & val  & 500     & 4{,}949  & --      & 500   & --   & --  & multi-hop (bridge/comp.) \\
MuSiQue-val        & val  & 500     & 8{,}225  & --      & 371   & 100  & 29  & multi-hop (2/3/4-hop) \\
\bottomrule
\end{tabular}
\caption{Dataset statistics. All splits are the ones distributed with the HippoRAG~2 suite. Hop counts for HotpotQA, 2Wiki, and LV-Eval are all 2-hop by construction; MuSiQue hop counts are derived from \texttt{len(question\_decomposition)}. NarrativeQA is single-document and not hop-structured.}
\label{tab:dataset_stats}
\end{table*}

\section{Tuning and Ablations}

\subsection{SHREK tuning}
\label{app:SHREK_tuning}

We tune the SHREK pipeline along several axes: the anchor pool size, the cross-encoder candidate pool, the final passage selection strategy, the hop expansion depth, and alias merging. A few hyperparameters are fixed from earlier diagnostic work and not tuned here. By diagnostic work we mean threshold sweeps run on the MuSiQue validation split before the tuning grid below, where we measured what fraction of gold and non-gold candidates each threshold keeps. These are the BFS bi-encoder threshold of $0.25$ (retains 77.6\% of gold passages while admitting only 0.2\% of non-gold), the direction threshold of $0.15$ (filters 97.4\% of all candidate triples while retaining 83.5\% of gold-leading triples), and the iterative bi-encoder prefilter of $0.15$.

We first widened the anchor pool and CE candidate pool, because per-stage recall analysis on the validation set showed a 9.3\% recall loss at the top-5 anchor cutoff for 2-hop MuSiQue questions. We then compared three final selection strategies: the unconstrained baseline (which returns 8 to 13 passages), chain-first selection (commit to the best anchor plus pair chain plus triple chain), and interleaved selection (best anchor, then alternate between unique hop-2 passages and other top anchors). Finally we tested the depth of the hop expansion, and whether to merge each entity with its synonym and buried-entity neighbours before traversing.

\begin{table*}[t]
\centering
\small
\begin{tabular}{llccc}
\toprule
Stage              & Configuration                & PopQA          & MuSiQue        & HotpotQA       \\
\midrule
Anchor pool top-5  & CE pool 20, unconstrained    & 0.766          & \textbf{0.507} & 0.735          \\
Anchor pool top-10 & CE pool 30, unconstrained    & \textbf{0.770} & 0.476          & \textbf{0.741} \\
\midrule
Chain-first        & top-5 passages               & 0.758          & 0.454          & 0.724          \\
Hedged             & top-5 passages               & \textbf{0.770} & 0.476          & 0.715          \\
Interleaved        & top-5 passages               & 0.767          & \textbf{0.483} & \textbf{0.736} \\
\midrule
Depth 2            & interleaved, merging on      & \textbf{0.774} & \textbf{0.513} & \textbf{0.745} \\
Depth 3            & interleaved, merging on      & 0.767          & 0.483          & 0.736          \\
Depth 4            & interleaved, merging on      & 0.769          & 0.510          & 0.743          \\
\midrule
Merging on         & depth 3, interleaved         & 0.767          & 0.483          & 0.736          \\
Merging off        & depth 3, interleaved         & \textbf{0.771} & \textbf{0.520} & \textbf{0.755} \\
\bottomrule
\end{tabular}%
\caption{SHREK tuning on the three validation sets. Token F1 with Llama-3.3-70B-Instruct as generator. Each stage holds earlier winners fixed and varies one component. Bold marks the best value per column within each stage, which is not always the row we keep. ``Merging on'' means each entity is merged with its synonym and buried-entity neighbours before traversing. CE is the cross-encoder.}
\label{tab:SHREK_tuning}
\end{table*}

The best configuration uses top-10 anchors drawn from a top-30 cross-encoder candidate pool, interleaved final selection, and alias merging off. Widening the anchor pool recovered the recall being lost at the top-5 cutoff, and it helped on PopQA and HotpotQA, but Token F1 on MuSiQue fell from $0.507$ to $0.476$ at this stage. We kept the wider pool because the extra anchors are what the later hop expansion extends from, and because MuSiQue recovers to $0.520$ once alias merging is turned off in the final stage, above the $0.507$ of the narrow pool. Interleaved selection beat chain-first and hedged selection because it puts the best anchor at position 1 then alternates between hop-2 passages and backup anchors, which hedges between single-hop and multi-hop questions within the 5-passage budget. The hop depth result was unexpected: depth 2 matches or beats depth 3 on all three datasets, and depth 4 is only slightly better than depth 3. The likely reason is that deeper hop expansion pulls in passages that pass the direction and bi-encoder filters but sit far from the question topic, and these compete for slots in the final 5-passage budget. We ran the depth sweep before the merging result, and the merging comparison was run at depth 3, so that is the depth we report. We did not test depth 2 with merging off, so the configuration we report may not be the best cell in this grid. The two depths are within $0.03$ Token F1 on every validation set. Turning alias merging off gave the single biggest improvement, especially on MuSiQue ($+3.7$) and HotpotQA ($+1.9$). Merging widens the frontier at every hop by pulling in other names and sub-parts of each entity, which fills the candidate pool with passages on related but different topics that push correct passages out of the top 5. On MuSiQue, merging is worse than having no synonym or buried-entity edges at all ($0.483$ against $0.507$ for the base graph), so these edges are useful to have in the graph but harmful to merge along. This ablation does not separate the effect of synonym edges from buried entity edges; isolating the two is left for future work. This is the configuration we report in \S\ref{sec:results}.

\subsection{Ablation tables}
\label{app:ablation_tables}

This subsection holds the tables for the graph enrichment, cross-encoder, and generator ablations discussed in \S\ref{sec:results}.

\begin{table}[ht]
  \centering
  \small
  \begin{tabular}{llccc}
    \toprule
    Pipeline & Graph & PopQA & MuSiQue & HotpotQA \\
    \midrule
    SHREK & Base    & 0.760          & 0.507          & 0.746          \\
    SHREK & Enriched & \textbf{0.771} & \textbf{0.520} & \textbf{0.755} \\
    \midrule
    Dijkstra & Base    & 0.752          & 0.402          & 0.650          \\
    Dijkstra & Enriched & \textbf{0.779} & \textbf{0.477} & \textbf{0.702} \\
    \midrule
    PCST & Base    & \textbf{0.773} & 0.495          & 0.704          \\
    PCST & Enriched & 0.766         & \textbf{0.515} & \textbf{0.724} \\
    \bottomrule
  \end{tabular}
  \caption{Graph enrichment ablation on validation sets. Token F1 with Llama-3.3-70B-Instruct as generator. Base~=~OpenIE triples only. Enriched~=~base + synonym, buried-entity, and related-to edges.}
  \label{tab:graph_enrichment}
\end{table}

\begin{table}[ht]
\centering
\small
\begin{tabular}{lccc}
\toprule
Scorer & PopQA & MuSiQue & HotpotQA \\
\midrule
Cross-encoder & \textbf{0.771} & \textbf{0.520} & \textbf{0.755} \\
Bi-encoder    & 0.030          & 0.159          & 0.260          \\
\bottomrule
\end{tabular}
\caption{Cross-encoder vs bi-encoder for concat scoring on validation sets. Token F1 with Llama-3.3-70B-Instruct as generator. Retrieval pipeline identical except for the scoring model used in anchor reranking and hop expansion.}
\label{tab:ce_ablation}
\end{table}

\begin{table}[ht]
\centering
\small
\begin{tabular}{lccc}
\toprule
Generator & PopQA & MuSiQue & HotpotQA \\
\midrule
Llama 3.3 70B & 0.771 & \textbf{0.520} & \textbf{0.755} \\
GPT-4o-mini   & \textbf{0.782} & 0.442          & 0.682          \\
\bottomrule
\end{tabular}
\caption{Generator comparison on validation sets. Token F1 with SHREK retrieval held fixed. Both generators read the validation prompt (App.~\ref{app:val_prompt}) and the top~$5$ retrieved passages, at temperature $0$, so this isolates the generator and not the prompt.}
\label{tab:generator}
\end{table}

\subsection{The synonym threshold}
\label{app:synonym}

The threshold of $0.88$ has no effect on the results we report. A synonym edge can only reach the retriever through alias merging (\S\ref{sec:kg_construction}), and we report SHREK with alias merging off, so no synonym edge is ever read. We confirmed this by building graphs at $0.80$ and $0.92$, which differ by $28$ times in synonym edge count.

With alias merging on, we sweep the threshold at $0.80$, $0.84$, $0.88$, $0.92$ and $0.95$ on all three validation sets, and read Recall@10. PopQA peaks at $0.88$ with $0.8375$, the highest of the five values, but a paired $t$-test cannot separate it from $0.84$ or $0.92$. At $0.80$ it falls to $0.725$, which is separable ($|t| = 3.36$). MuSiQue and HotpotQA show the same shape. The values at $0.84$, $0.88$ and $0.92$ sit close together, none of them separable from merging off ($|t| \leq 1.40$), and HotpotQA runs between $0.988$ and $1.000$ across every setting, so on that set the metric has no room to move. The one value that is clearly worse is $0.80$. So no threshold is proven best. There is a range from about $0.84$ to $0.92$ where the results are the same, and we take $0.88$ because it scored highest on PopQA and sits inside that range.

Alias merging still does not beat leaving it off on any validation set, and the denser the synonym graph is the more it costs. On MuSiQue, merging at $0.80$ takes $2.7$ times the retrieval time of merging off, and on HotpotQA it takes $2.5$ times, in both cases with no recall gain. Retrieval time falls as the threshold rises on all three sets.

Edge quality remains poor even at $0.88$. Lowering the threshold mostly adds more bad edges rather than good ones, with failures such as \texttt{25 august} joined to \texttt{august 20} and \texttt{525} joined to \texttt{572}. A filter that refuses to join two numeric strings would remove most of these errors. We have not implemented it.

\subsection{The direction filter}
\label{app:direction}

We test what the direction filter contributes by lowering $\theta_{\text{dir}}$ until every candidate triple passes, so hop expansion goes by connectivity alone. We run both settings on MuSiQue validation questions. Recall@1 does not change. Recall@5 falls by $0.0034$, Recall@20 by $0.0150$, and Recall@10 by $0.0200$, the largest of the four. Mean retrieval time rises by $40\%$ on the same hardware. The filter is doing cost work rather than quality work. Without it the retriever reaches roughly the same passages, it just reads more triples on the way. At $\theta_{\text{dir}} = 0.15$ it drops $97.4\%$ of candidate triples while keeping $83.5\%$ of the triples that lead to a gold passage (App.~\ref{app:SHREK_tuning}), which is where the time saving comes from. We keep the filter on because the accuracy difference is small and the time difference is not. 

\subsection{BERTScore}
\label{app:bertscore}

Table~\ref{tab:bertscore} reports BERTScore~F1 on the test splits for our three pipelines. We only report BERTScore for the main results table; we do not report it for ablations to keep the main text focused on Token~F1.

\begin{table}[ht]
\centering
\small
\begin{tabular}{lccc}
\toprule
Dataset & Dijkstra & PCST & SHREK \\
\midrule
NQ             & 0.914 & 0.917 & 0.915 \\
PopQA          & 0.916 & 0.917 & 0.918 \\
MuSiQue        & 0.885 & 0.901 & 0.902 \\
2Wiki          & 0.889 & 0.909 & 0.918 \\
HotpotQA       & 0.911 & 0.932 & 0.935 \\
LV-Eval        & 0.857 & 0.852 & 0.859 \\
NarrativeQA    & 0.880 & 0.885 & 0.882 \\
\bottomrule
\end{tabular}
\caption{BERTScore~F1 on the test splits. All pipelines use NV-Embed-v2 embeddings, GPT-4o-mini as generator, and the top $5$ retrieved passages as context.}
\label{tab:bertscore}
\end{table}

The BERTScore ranking matches the Token~F1 ranking in Table~\ref{tab:all_benchmarks}: SHREK is best overall, PCST is close behind, and Dijkstra trails. The absolute BERTScore values are high across all pipelines because BERTScore measures contextual similarity rather than exact token overlap, so semantically close answers (``USA'' vs ``United States'', or short vs more elaborated forms of the same entity) get credit that Token~F1 does not.

\section{Diagnostics and Tests}

\subsection{Cross-encoder discrimination by hop count}
\label{app:discrimination}

The failure analysis in \S\ref{sec:discussion} rests on one measurement, so we describe it here. For each MuSiQue validation question we build the all-gold chain, which concatenates every gold supporting passage, and the best distractor chain, which replaces one gold passage with the highest-scoring non-gold passage in the BFS pool. We define the discrimination gap as the cross-encoder score of the all-gold chain minus the cross-encoder score of the best distractor chain, averaged over questions. A positive gap means the cross-encoder prefers the gold chain.

The gap is $+0.042$ at 2-hop ($n = 371$), $-0.080$ at 3-hop ($n = 100$), and $-0.100$ at 4-hop ($n = 29$). At 2 hops the cross-encoder prefers the gold chain. At 3 and 4 hops it prefers a chain with one distractor in it more often than not. This is the same two-passage comparison at every depth, so what changes is the question and not the measurement: as the chain the question needs gets longer, the cross-encoder gets worse at preferring a real second gold passage over a distractor on the same topic. At 3 and 4 hops the gold side is a cut-down gold set, since those questions have more than two gold passages. The 4-hop cell rests on $29$ questions, so we lean on the 3-hop cell. This is the measurement behind the 2-hop ceiling we report in the Limitations.

\subsection{Significance tests}
\label{app:significance}

The markers in Tables~\ref{tab:all_benchmarks} and~\ref{tab:recall} come from a paired bootstrap, and the McNemar column in Table~\ref{tab:buckets} from a second test, so we describe both here. SHREK and PCST answer the same questions, so we compare them with a paired bootstrap. For each test set we draw $N$ questions with replacement, where $N$ is the size of that set, and take the mean score of SHREK minus the mean score of PCST. We repeat this $10{,}000$ times with a fixed seed. The two-sided $p$-value is twice the smaller of two fractions, the fraction of samples at or below zero and the fraction at or above it, and the 95\% interval runs from the 2.5th to the 97.5th percentile. We then apply a Holm correction across datasets, separately for Token F1 (seven sets) and Recall@5 (five sets). Recall@5 uses the gold definition from Table~\ref{tab:recall}. \citet{gutierrez2025rag} also use a bootstrapped test, against their dense baseline. For SHREK against HippoRAG~2 we use an exact McNemar test on our rerun instead (App.~\ref{app:agreement}), because the reported numbers come with no per-question outputs. Table~\ref{tab:significance} gives the bootstrap results.

\begin{table*}[t]
\centering
\small
\begin{tabular}{llrrrrr}
\toprule
Metric & Dataset & SHREK & PCST & Diff & 95\% CI & $p_{\text{Holm}}$ \\
\midrule
Token F1 & NQ          & 59.36 & 60.83 & $-1.47$  & $[-3.32, 0.39]$    & 0.595 \\
         & PopQA       & 62.32 & 62.14 & $+0.17$  & $[-1.20, 1.59]$    & 0.860 \\
         & MuSiQue     & 45.77 & 44.63 & $+1.14$  & $[-1.30, 3.57]$    & 0.860 \\
         & 2Wiki       & 60.01 & 55.39 & $+4.62$  & $[1.73, 7.48]$     & \textbf{0.013} \\
         & HotpotQA    & 73.01 & 70.56 & $+2.46$  & $[0.72, 4.16]$     & \textbf{0.024} \\
         & LV-Eval     & 12.22 &  9.24 & $+2.99$  & $[-0.90, 7.06]$    & 0.595 \\
         & NarrativeQA & 24.48 & 26.11 & $-1.63$  & $[-4.60, 1.41]$    & 0.860 \\
\midrule
Recall@5 & NQ          & 63.21 & 75.11 & $-11.90$ & $[-13.99, -9.88]$  & $\mathbf{<10^{-4}}$ \\
         & PopQA       & 53.90 & 50.25 & $+3.65$  & $[2.55, 4.75]$     & $\mathbf{<10^{-4}}$ \\
         & MuSiQue     & 64.41 & 67.18 & $-2.78$  & $[-4.54, -1.06]$   & \textbf{0.001} \\
         & 2Wiki       & 80.65 & 73.80 & $+6.85$  & $[5.30, 8.45]$     & $\mathbf{<10^{-4}}$ \\
         & HotpotQA    & 92.75 & 91.50 & $+1.25$  & $[-0.10, 2.60]$    & 0.080 \\
\bottomrule
\end{tabular}
\caption{Paired bootstrap tests of SHREK against PCST on the test sets. Means are on a $0$ to $100$ scale, and Diff is SHREK minus PCST. $p_{\text{Holm}}$ is Holm-corrected across datasets within each metric, and bold marks $p_{\text{Holm}} < 0.05$. $<10^{-4}$ means no bootstrap sample crossed zero.}
\label{tab:significance}
\end{table*}

\subsection{Question-level agreement}
\label{app:agreement}

Table~\ref{tab:buckets} breaks down, per dataset, how often SHREK and HippoRAG~2 agree, and which system uniquely solves each question.
\begin{table*}[t]
\centering
\small
\begin{tabular}{lccccccc}
\toprule
Dataset & N & Both correct & HippoRAG~2 only & SHREK only & Both wrong & McNemar $p_{\text{Holm}}$ \\
\midrule
PopQA          & 1000 & 54.5\% & 3.3\%  & \textbf{9.2\%}  & 33.0\% & $<0.001^\ast$ \\
NQ      & 1000 & 53.9\% & \textbf{8.5\%}  & 6.5\%  & 31.1\% & 0.36 \\
HotpotQA       & 1000 & 69.3\% & 6.2\%  & \textbf{6.8\%}  & 17.7\% & 1.00 \\
2WikiMultihopQA& 1000 & 57.4\% & \textbf{15.7\%} & 4.6\%  & 22.3\% & $<0.001^\ast$ \\
MuSiQue        & 1000 & 38.1\% & \textbf{12.6\%} & 9.0\%  & 40.3\% & 0.068 \\
NarrativeQA    & 293  & 11.6\% & \textbf{7.5\%}  & 5.8\%  & 75.1\% & 1.00 \\
\bottomrule
\end{tabular}
\caption{Question-level agreement between SHREK and our rerun of HippoRAG~2 on six test sets. We did not rerun LV-Eval. Both systems use GPT-4o-mini as generator and the top $5$ retrieved passages as context. A question is counted correct if Token F1 $\geq 0.5$. Bold marks the larger ``only'' group per dataset. The last column is an exact McNemar test on the two ``only'' groups, Holm-corrected across the six datasets, and $^\ast$ marks $p_{\text{Holm}} < 0.05$.}
\label{tab:buckets}
\end{table*}

\section{Prompts}
\label{app:prompts}

This appendix lists every prompt used in our pipelines. The same prompts are used across all three pipelines unless noted.

\subsection{NER Prompt (query-side)}
\label{app:ner_prompt}

Used in the NER + cosine + LCS query matcher (\S\ref{sec:matchers}) to extract candidate entity strings from the question. The model is GPT-4o-mini at temperature~$0$.

\begin{quote}
\small
\ttfamily
You are a highly effective entity extraction system.\\[4pt]Extract all named entities that are important for solving the question. Only extract entities that would have a Wikipedia article or a node in a knowledge base --- specific people, places, organizations, events, works, theories, or concepts with a proper name. Do NOT extract generic nouns.\\[4pt]Output valid JSON only, no extra text:\\\{"named\_entities": ["Entity1", "Entity2"]\}\\[4pt]Example:\\[2pt]Question: Which magazine was started first Arthur's Magazine or First for Women?\\\{"named\_entities": ["Arthur's Magazine", "First for Women"]\}
\end{quote}

\subsection{Triple Filter Prompt (query-side)}
\label{app:triple_prompt}

Used in the triple-matching + LLM filter query matcher (\S\ref{sec:matchers}) by the Dijkstra and (optionally) PCST pipelines. The system message and few-shot demonstrations are copied from HippoRAG~\citep{gutierrez2024hipporag} so that our setup is comparable. The model is GPT-4o-mini at temperature~$0$.

\begin{quote}
\small
\ttfamily
You are a critical component of a high-stakes question-answering system used by top researchers and decision-makers worldwide. Your task is to filter facts based on their relevance to a given query, ensuring that the most crucial information is presented to these stakeholders. The query requires careful analysis and possibly multi-hop reasoning to connect different pieces of information. You must select up to 4 relevant facts from the provided candidate list that have a strong connection to the query, aiding in reasoning and providing an accurate answer. The output should be in JSON format, e.g., \{"fact": [["s1", "p1", "o1"], ["s2", "p2", "o2"]]\}, and if no facts are relevant, return an empty list, \{"fact": []\}. The accuracy of your response is paramount, as it will directly impact the decisions made by these high-level stakeholders. You must only use facts from the candidate list and not generate new facts.
\end{quote}

The prompt is paired with 7 few-shot examples from HippoRAG's public prompt set; see their repository for the complete list.

\subsection{Generator Prompt (test set)}
\label{app:test_prompt}

Used for all main test-set results (\S\ref{sec:results}). This is the chain-of-thought prompt from HippoRAG~2~\citep{gutierrez2025rag} with a single worked example. The generator is GPT-4o-mini at temperature~$0$.

\begin{quote}
\small
\ttfamily
As an advanced reading comprehension assistant, your task is to analyze text passages and corresponding questions meticulously. Your response start after ``Thought: '', where you will methodically break down the reasoning process, illustrating how you arrive at conclusions. Conclude with ``Answer: '' to present a concise, definitive response, devoid of additional elaborations.\\[4pt]Example:\\[2pt]Wikipedia Title: The Last Horse\\The Last Horse (Spanish: El último caballo) is a 1950 Spanish comedy film directed by Edgar Neville starring Fernando Fernán Gómez.\\[2pt] Wikipedia Title: Neville A. Stanton\\Neville A. Stanton is a British Professor of Human Factors and Ergonomics at the University of Southampton...\\[4pt]Question: When was Neville A. Stanton's employer founded?\\Thought: The employer of Neville A. Stanton is University of Southampton. The University of Southampton was founded in 1862.\\ Answer: 1862.\\[4pt] \{context\}
\end{quote}

The full HippoRAG CoT prompt has five supporting passages in the demonstration; we kept the same format to match their reporting setup exactly.

\subsection{Generator Prompt (validation set)}
\label{app:val_prompt}

Used for all validation-set tuning with Llama-3.3-70B-Instruct at temperature~$0$.

\begin{quote}
\small
\ttfamily
You are an advanced question answering assistant. Your task is to answer the question accurately.\\[4pt]Guidelines:\\
- If the answer is in the provided context then use it 
otherwise use your own knowledge.\\
- Provide concise, direct answers --- typically just the key 
information (a name, date, occupation, or short phrase).\\
- If the question asks about ONE person/thing, provide ONE 
answer.\\
- If the question asks about multiple items, list all relevant 
answers.\\
- Only provide longer explanations if the question explicitly 
asks for description or explanation.\\
- Conclude with your answer presenting a concise, definitive 
response.\\[4pt]
Examples:\\
Question: What is Marie Curie's occupation?\\
Answer: Physicist\\[2pt]
Question: When was the Eiffel Tower completed?\\
Answer: 1889\\[2pt]
Question: Who wrote Romeo and Juliet?\\
Answer: William Shakespeare\\[2pt]
Question: Which countries border France?\\
Answer: Belgium, Luxembourg, Germany, Switzerland, Italy, Spain\\[4pt]
Context:\\
\{context\}
\end{quote}

The validation prompt is simpler than the test prompt because we did not re-run tuning under the HippoRAG CoT format after switching to GPT-4o-mini for the final test-set results.

\subsection{NV-Embed-v2 Query Instruction}
\label{app:nv_embed_instruction}

NV-Embed-v2~\citep{2405.17428v3} uses asymmetric retrieval: queries are encoded with a task instruction prefix, while passages, entities, and triples are encoded without one. The instruction we use is the default retrieval instruction from the model card:

\begin{quote}
\small
\ttfamily
Instruct: Given a question, retrieve passages that answer the 
question\\
Query: \{question\}
\end{quote}

Passage, entity, and triple embeddings are computed over the raw text with no instruction prepended.                     

\end{document}